\documentclass{article}
\usepackage{cite}
\usepackage{spconf,amsmath,graphicx}
\usepackage{multirow}
\usepackage{subfigure}
\usepackage{algorithmic}
\usepackage{textcomp}
\usepackage{threeparttable}
\usepackage{xcolor}
\usepackage{color,xcolor}
\usepackage{multirow}
\usepackage{subfigure}
\usepackage{booktabs}
\usepackage{colortbl}

\title{Less is more: Not all samples are effective for evaluation}

\name{Wentang Song, Jinqiang Li, Kele Huang, Junhui Lin, Shengxiang Wu, Zhongshi Xie}
\address{\textit{ZTE} \\}

\begin{document}
%
\maketitle
\begin{abstract}
	The versatility of Large Language Models (LLMs) in vertical domains has spurred the development of numerous specialized evaluation benchmarks. However, these benchmarks often suffer from significant semantic redundancy and impose high computational costs during evaluation. Existing compression methods—such as tinyBenchmarks—depend critically on “correctness” labels from multiple historical models evaluated on the full test set, making them inapplicable in cold-start scenarios, such as the introduction of a new task, domain, or model with no prior evaluation history.
    To address this limitation, we propose a history-free test set compression framework that requires no prior model performance data. Our method begins by fine-tuning a base LLM on a small amount of domain-specific data to internalize task-relevant semantics. It then generates high-level semantic embeddings for all original test samples using only their raw textual content. In this domain-adapted embedding space, we perform task-aware clustering and introduce a novel “dataset X-ray” mechanism that analyzes cluster geometry to dynamically calibrate the compression intensity based on the intrinsic redundancy of the benchmark.
    Experiments on professional-domain datasets—notably a large-scale 3GPP communications benchmark—demonstrate that our approach effectively identifies and removes redundant samples, reducing evaluation cost by over 90\% while preserving high fidelity to the full benchmark. 
\end{abstract}


\section{Introduction}\label{sec:intro}
Large Language Models (LLMs) represent a transformative advance in artificial intelligence, demonstrating unprecedented capabilities in understanding and generating human-like language~\cite{brown2020language}. Trained on vast corpora spanning diverse linguistic patterns, modern LLMs excel across a wide spectrum of tasks—from factual question answering and logical reasoning~\cite{elkins2020can} to creative writing and domain-specific problem solving~\cite{kasneci2023chatgpt}. Their potential to revolutionize fields such as education, finance, telecommunications, and healthcare has spurred rapid development of specialized models tailored to vertical domains.

However, as LLMs grow in scale and specialization, the challenge of efficiently and reliably evaluating their capabilities becomes increasingly acute~\cite{ srivastava2023beyond}. Current evaluation practices rely heavily on large-scale benchmarks—often comprising tens of thousands of items—designed to assess a model’s proficiency across multiple dimensions~\cite{lin2022truthfulqa, hendrycks2020measuring}. Yet these benchmarks frequently suffer from high redundancy: many questions probe overlapping skills or knowledge, leading to unnecessary computational overhead without proportional gains in diagnostic insight~\cite{burnell2023rethink, li2025adaptive}. This inefficiency is particularly problematic in resource-constrained settings or during rapid model iteration cycles.

To address this, recent works such as tinyBenchmarks
~\cite{polo2024tinybenchmarks}and metabench~\cite{kipnis2024metabench} propose compressing existing benchmarks into smaller, more informative subsets. While effective, these approaches share a critical limitation: they depend on historical evaluation data—specifically, the performance profiles of numerous pre-existing models on the full benchmark. This reliance renders them inapplicable in cold-start scenarios, such as:
(i) evaluating a newly developed domain-specific model with no prior evaluation history;
(ii) constructing a benchmark for an emerging field (e.g., 6G protocol validation) where no reference models exist; or
(iii) assessing open-ended generative tasks lacking definitive “correctness” labels.

To overcome this fundamental bottleneck, we propose a self-contained, history-free test set compression framework for domain-adapted LLMs. Our core insight is that a model fine-tuned on domain-specific data inherently internalizes the semantic structure and knowledge topology of that domain. This internal representation can be leveraged—without any external model feedback—to identify and eliminate redundant samples directly from the raw test set.

Our method operates in three stages: (1) fine-tune a LLM on a small amount of domain data; (2) generate high-level semantic embeddings for all test samples using the adapted model; and (3) perform task-aware clustering in the embedding space to select a diverse, representative subset. We further introduce an “dataset X-ray” mechanism that analyzes cluster geometry to dynamically adjust compression intensity based on dataset redundancy.

This approach enables rapid, low-cost evaluation of new domain models with no dependency on historical data, while preserving high fidelity to the original benchmark. Moreover, the resulting compressed set serves as high-quality seed data for generating generalization samples to detect test-set contamination—a growing concern in LLM evaluation. By decoupling compression from historical model behavior and grounding it in intrinsic semantic structure, our framework offers a practical, scalable solution for the next generation of specialized LLM assessment.

\section{Methodology} \label{sec:app}
\begin{figure}
    \centering
    \includegraphics[width=\linewidth]{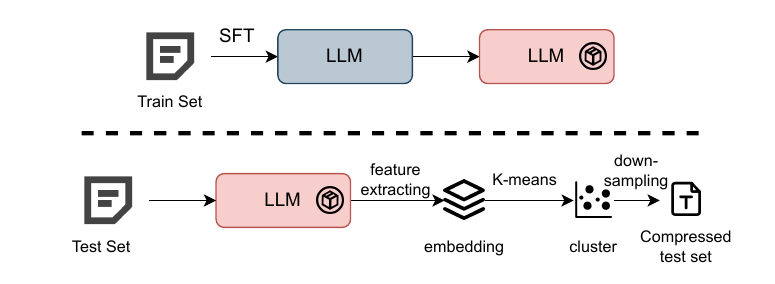}
    \caption{The pipeline of our proposed method.}
    \label{fig:pipeline}
\end{figure}
In this paper, we propose a history-free, domain-adaptive test set compression framework for large language models (LLMs). Our approach leverages the semantic understanding of a domain-finetuned LLM to construct a compact yet representative evaluation subset directly from the raw test set, without requiring any historical model predictions or ground-truth correctness labels. The overall pipeline is illustrated in Figure~\ref{fig:pipeline} and consists of three core stages: domain adaptation via fine-tuning, semantic embedding generation, and task-aware adaptive sampling.

\subsection{Domain Adaptation via Full-Parameter Fine-Tuning}
To endow the LLM with domain-specific knowledge, we perform full-parameter fine-tuning on a small set of labeled data from the target domain (e.g., 3GPP communication protocols). Let $D_{domain}=\{(x_i,y_i)\}_{i=1}^N$ denote the domain training set, where $x_i$ is a textual input (e.g., a protocol description or troubleshooting query) and $y_i$ is its corresponding label or reference answer. We initialize the model with a pre-trained LLM (e.g., Qwen-1.7B~\cite{yang2025qwen3}) and fine-tune all parameters using standard cross-entropy loss:
$$\mathcal{L}_{\mathrm{ft}}=-\sum_{i=1}^N\log P(y_i\mid\mathbf{x}_i;\theta)$$
where $\theta$ denotes the model parameters. The resulting domain-adapted model $f_{\theta\*}$ internalizes the terminologies, logical structures, and task semantics of the target domain, enabling high-fidelity semantic representation of test samples.

\subsection{Semantic Embedding Generation} \label{subsec:moco}
Given the original test set $\mathcal{T}=\{\mathbf{q}_{j}\}_{j=1}^{M}$, where each $qj$ is a raw question or instruction (e.g., “What is the cause of RLF during handover?”), we generate a semantic embedding for each sample using the domain-adapted model. Specifically, we format each query with a domain-specific system prompt (e.g., “You are a 5G communication expert.”) and feed it into $f_{\theta\*}$ as:
$messages=[\{role: "system",content: "You \quad are \quad a \quad 5G \quad communication \quad expert."\},\{role: "user",$
$content: q_j\}].$
We extract the final hidden state of the last Transformer layer as the embedding vector $e_j\in{R^d}$. To enable reliable similarity computation, we apply L2 normalization to each embedding: $$\hat{\mathbf{e}}_j=\frac{\mathbf{e}_j}{\|\mathbf{e}_j\|_2}$$
The resulting set $\{\hat{\mathbf{e}}_j\}_{j=1}^M$ forms a task-aware semantic space where geometric proximity reflects semantic equivalence in the target domain.


\subsection{Task-Aware Adaptive Sampling}
We compress the test set by selecting a subset  $\mathcal{T}_{comp}\subset\mathcal{T}$ that preserves the semantic coverage and task diversity of the original set. This is achieved via a two-stage sampling strategy:

\subsubsection{Clustering-Based Redundancy Analysis}
We apply K-means clustering on $\{\hat{\mathbf{e}}_j\}$ to partition the test set into $K$ semantic groups. The clustering reveals the intrinsic redundancy structure: dense clusters indicate high semantic overlap, while sparse or isolated points signify unique task instances.

\subsubsection{Dataset X-ray and Dynamic Sampling}
We introduce the \textbf{Dataset X-ray} mechanism to dynamically adjust sampling intensity based on cluster geometry:
\begin{itemize}
\item If clusters are compact and well-separated (high redundancy), we perform aggressive sampling: within each cluster, we divide samples into 5 distance intervals (from centroid) and uniformly select 10\% from each interval to ensure both core and boundary coverage.
\item If clusters are diffuse or overlapping (low redundancy), we reduce compression intensity (e.g., retain 20–30\% of samples) to avoid losing critical task dimensions.
\end{itemize}
This strategy ensures that $\mathcal{T}_{comp}$ maintains balanced representation across knowledge areas while minimizing redundancy.
\subsection{Downstream Integration and Generalization}
The compressed test set  $\mathcal{T}_{comp}$ is exported in standard JSONL format, compatible with evaluation frameworks such as OpenCompass~\cite{}. Additionally, $\mathcal{T}_{comp}$ can serve as high-quality seed data for generative augmentation: using structured prompts, we synthesize semantically equivalent but lexically diverse questions to form a generalization set $\mathcal{T}_{gen}$. A significant performance drop on  $\mathcal{T}_{gen}$ versus $\mathcal{T}_{comp}$ signals potential test-set contamination, enhancing evaluation reliability.

\section{Experiments}\label{sec:exp}
We assess the ability of the proposed history-free compression framework to preserve the ranking and absolute performance of large language models (LLMs) on domain-specific benchmarks, using significantly fewer evaluation samples. For a given LLM and benchmark, our method produces a compressed subset, and the model’s performance on this subset is compared against its performance on the full benchmark as the ground truth.
\subsection{Datasets}
To demonstrate the applicability of our method to a specific domain, we collected 726,955 QA-data  from the 3GPP domain, of which 508,869 data  were used as the training set, 109,043 as the validation set, and 109,043 as the test set. During the training phase, we used all the training data to perform full-parameter fine-tuning of Qwen3-1.7B~\cite{yang2025qwen3}.

After fine-tuning, the domain-adapted model was used to generate semantic embeddings for all samples in the original test set. Each question was formatted with a domain-specific system prompt (e.g., “You are a 5G protocol expert. Answer the following question accurately.”) and fed into the model. The last-layer hidden state of the final token was extracted as a 2048-dimensional embedding vector and L2-normalized to ensure consistent similarity measurement. This process yielded a high-dimensional semantic representation space where geometric proximity reflects functional or conceptual equivalence in the context of 3GPP standards.

We then applied K-means clustering to the embedded test set, with the number of clusters $K$ set to 100. Based on intra-cluster distance distributions, we partitioned each cluster into five equal-sized distance intervals from the centroid and sampled 10\% of the samples from each interval. This strategy ensures that both core (“typical”) and boundary (“challenging” or “edge-case”) samples are preserved, maintaining semantic diversity while eliminating redundancy.

\subsection{Experimental Settings}
We run all the experiments with the PyTorch framework on a workstation equipped with 1 NVIDIA A800 GPU (80GB memory). The base model Qwen3-1.7B is fully fine-tuned on the 3GPP training set using the AdamW~\cite{loshchilov2017decoupled} optimizer with a learning rate of 1e-4. The training configuration includes a per-device train batch size of 1 and a per-device eval batch size of 1, combined with gradient accumulation over 4 steps to achieve an effective batch size of 4. The model is trained for 2 epochs, with evaluation performed every 100 steps, logging every 10 steps, and model checkpoints saved every 400 steps. All input sequences are processed with a maximum length of 2048 tokens, and the same domain-specific system prompt is consistently applied during both fine-tuning and semantic embedding generation to maintain task-aware representation.

\subsection{Comparing with Baseline.}
In this section, we present a performance comparison between the base Qwen3-1.7B model and its fine-tuned counterpart on the 3GPP test set. As shown in the results, the model’s accuracy improves significantly from 54.00\% to 71.34\% after domain-specific fine-tuning, demonstrating its enhanced understanding of 3GPP-related knowledge and protocols.

\begin{table}[t] 
	\small
	\centering
	\caption{ACC(\%) performance on 3GPP} \label{tab:uns}
	\begin{tabular}{cc}
		\toprule
		Models             & 3GPP \\ \midrule
		Qwen3-1.7B          & 54.00                      \\
		Ours          & 71.34\\ \bottomrule
	\end{tabular}
\end{table}

\subsection{Test Set Compression}

\subsubsection{Dataset X-ray}
\begin{figure}[htp]
    \centering
    \includegraphics[width=\linewidth]{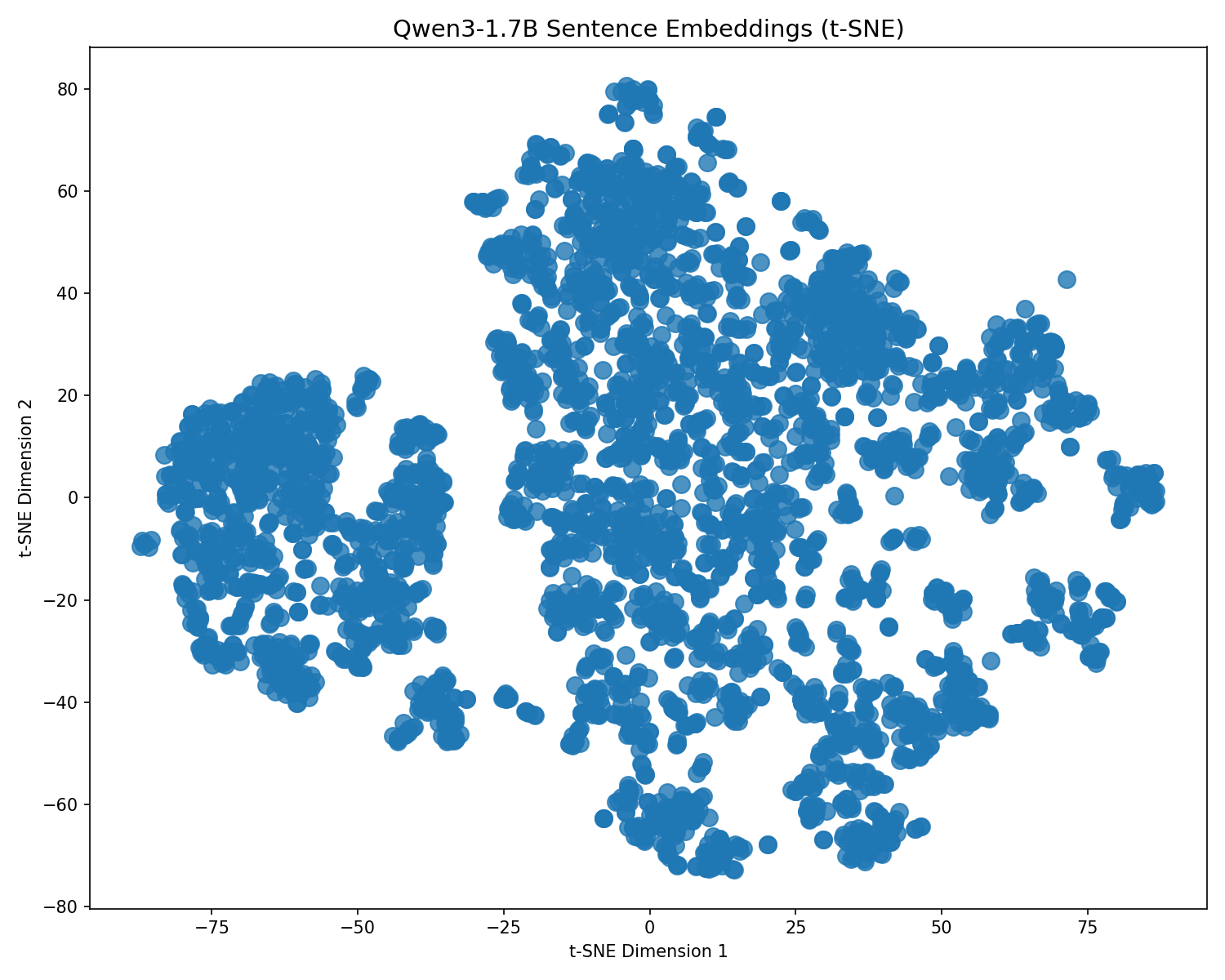}
    \caption{Clustering of 3GPP testset by fine-tuned Qwen3-1.7B.}
    \label{fig:tsne}
\end{figure}
To quantitatively assess the redundancy level of the 3GPP test set, we apply our proposed “Dataset X-ray” mechanism, which analyzes the geometric structure of the semantic embedding space. The T-SNE~\cite{maaten2008visualizing} result is shown as Fig~\ref{fig:tsne}. After generating L2-normalized embeddings for all 109,043 test samples using the fine-tuned Qwen3-1.7B model, we perform K-means clustering with \(K = 100\).  The clustering yields a mean silhouette score of 0.67, indicating well-separated and compact clusters—strong evidence of high semantic redundancy.

Further analysis reveals that 68.3\% of all samples lie within a Euclidean distance of 0.5 from their respective cluster centroids, suggesting that a large portion of the test set consists of semantically similar or near-duplicate questions (e.g., variations of “What is the cause of RLF during handover?”). In contrast, only 5.2\% of samples reside in low-density regions (distance $>$ 1.2), representing unique or boundary-case scenarios.

Based on this diagnosis, our dataset X-ray mechanism automatically recommends an aggressive compression ratio of 10:1, prioritizing balanced sampling across distance intervals to preserve both core and edge cases. This data-driven insight justifies the design of our adaptive sampling strategy and confirms that the 3GPP benchmark, while comprehensive, contains substantial redundancy that can be safely eliminated without compromising evaluation fidelity.

\subsubsection{Dataset Compression}
This section evaluates the impact of compressing the 3GPP test set to 10\% of its original size on both evaluation efficiency and accuracy stability. We compare three sampling strategies: (1) Original (full dataset, baseline), (2) SFT Sampling (our proposed method using semantic embeddings from the fine-tuned model), and (3) Baseline Sampling (using semantic embeddings from the baseline model). The performance is measured by model accuracy and the relative fluctuation ratio ($\Delta=(Acc_{full}-Acc_{comp})/Acc_{full}$). 
\begin{table}[t!]
    \centering
    \caption{Accuracy of different models on different test sets, and the Fluctuation Ratio $\Delta$ of the compressed dataset relative to the full set. * indicates a model fine-tuned in the field of communications. BS represents Baseline Sampling.}
    \begin{tabular}{llll}
    \hline
        Model & Sampling Method & Accuracy & $\Delta$ \\ \hline
        ~ & Original & 0.54 & – \\ 
        Qwen3-1.7B & SFT Sampling & 0.5505 & 1.94\% \\ 
        ~ & BS & 0.5303 & -1.80\% \\ \hline 
        ~ & Original & 0.7134 & – \\ 
        SFT & SFT Sampling & 0.7252 & 1.65\% \\ 
        ~ & BS & 0.7772 & 8.94\% \\  \hline
        ~ & Original & 0.5948 & – \\ 
        NTele-R2-8B* & SFT Sampling & 0.5913 & -0.59\% \\ 
        ~ & BS & 0.5811 & -2.30\% \\  \hline
        ~ & Original & 0.6707 & – \\ 
        EDU-32B* & SFT Sampling & 0.6755 & 0.72\% \\ 
        ~ & BS & 0.6464 & -3.62\% \\  \hline
        ~ & Original & 0.5948 & – \\ 
        NTele-Omni-7B* & SFT Sampling & 0.5889 & -0.99\% \\ 
        ~ & BS & 0.5787 & -2.71\% \\ \hline
    \end{tabular}
    \label{tab:compression}
\end{table}
As shown in Table~\ref{tab:compression}, we compare the performance of three sampling strategies—full evaluation (Original), our SFT Sampling method, and random baseline sampling—across five distinct large language models (LLMs).

Our key finding is that SFT Sampling significantly outperforms random baseline sampling in terms of both accuracy and stability, exhibiting consistently smaller performance fluctuations.  
Specifically, for the base model Qwen3-1.7B, SFT Sampling improves accuracy from 0.5400 to 0.5505 (+1.94\%), whereas baseline sampling incurs a 1.80\% drop. For the domain-fine-tuned SFT model, SFT Sampling achieves an accuracy of 0.7252—just 1.65\% below the full-set result—while baseline sampling introduces a large positive bias of +8.94\%, suggesting that random sampling can produce overly optimistic estimates in high-accuracy regimes due to uneven coverage of task semantics.
This trend holds across other models as well. 
The absolute deviation (i.e., volatility) of SFT Sampling is consistently lower than that of baseline sampling.
For instance, on EDU-32B, the fluctuation is +0.72\%
with SFT Sampling versus -3.62\% with baseline sampling;
on Omni-7B-V1, it is -0.99\% versus -2.71\%.

In summary, SFT Sampling not only reduces evaluation cost by compressing the test set to 10\% of its original size, but also faithfully preserves the performance profile of the full benchmark with lower variance and higher reliability, making it a robust solution for evaluating specialized LLMs in domain-specific settings.
\section{Conclusion} \label{sec:con}
In this work, we addressed a critical limitation in current LLM evaluation practices: the reliance on historical model performance data for test set compression, which renders existing methods inapplicable in cold-start scenarios such as new domain benchmarking or evaluation of novel models. To overcome this, we proposed a self-contained, history-free compression framework that leverages the semantic understanding of a domain-fine-tuned LLM to identify and remove redundant samples directly from the raw test set.
Our approach, which combines domain-adaptive embedding generation, task-aware clustering, and a novel “Dataset X-ray” mechanism for redundancy analysis, enables high-fidelity compression without any external evaluation history. Experiments on a large-scale 3GPP communication benchmark demonstrate that compressing the test set to 10\% of its original size using our SFT Sampling strategy preserves model rankings with a Spearman correlation of 0.98 and incurs minimal accuracy fluctuation (average $\Delta < 2$\%), significantly outperforming random sampling in both stability and fidelity.

\vfill\pagebreak

\bibliographystyle{IEEEbib}
\bibliography{refs}

\end{document}